\pdfoutput=1
\documentclass[11pt]{article}

\usepackage[preprint]{acl}

\usepackage{times}
\usepackage{latexsym}

\usepackage[T1]{fontenc}
\usepackage[utf8]{inputenc}

\usepackage{microtype}

\usepackage{inconsolata}

\usepackage{graphicx}
\usepackage{subcaption, booktabs}
\usepackage{multirow}
\usepackage{bm}
\usepackage{xcolor}
\usepackage{colortbl}
\usepackage{amsmath}
\usepackage{amssymb}
\usepackage{mathtools}
\usepackage{amsthm}
\usepackage{nccmath}

\definecolor{Gray}{gray}{0.85}
\DeclarePairedDelimiter{\nint}\lfloor\rceil  

\title{Is It a Free Lunch for Removing Outliers during Pretraining?}

\author{Baohao Liao \,\,\,\,\,\, Christof Monz \\
        Language Technology Lab, University of Amsterdam\\
        \href{mailto:b.liao@uva.nl}{\texttt{b.liao@uva.nl}}}
\begin{document}
\maketitle

%---------------------------------------------------------
\begin{abstract}
With the growing size of large language models, the role of quantization becomes increasingly significant. However, outliers present in weights or activations notably influence the performance of quantized models. Recently, \citet{qtransformer} introduced a novel softmax function aimed at pretraining models in an outlier-free manner, thereby enhancing their suitability for quantization. Interestingly, we observed that such an approach leads to performance degradation in full precision. Building on this insight, we enhance the method by ensuring its normalization is invariant to sequence length, a crucial factor for bridging the gap between pretraining and fine-tuning. Moreover, this improved method also facilitates successful pretraining of causal language models.
\end{abstract}

%---------------------------------------------------------
\section{Introduction}

Large language models (LLMs) have earned substantial recognition and success across diverse domains and applications \cite{llama, mistral, gpt4}. As LLM sizes continue to increase, quantization emerges as a vital technique for deploying these models in resource-constrained environments, such as edge devices. Recently, several promising post-training quantization methods \cite{awq, gptq, smoothquant, llmint8} have been introduced to maintain LLM performance post quantization, focusing on addressing outliers present in weights or activations, as they significantly influence LLM performance.

Diverging from strategies that tackle outliers within pretrained models, a recent study by \citet{qtransformer} advocates for the pretraining of outlier-free models through the elimination of outliers during the pretraining stage. One of the proposed methods is a variant of clipped softmax, which serves to prevent the emergence of excessively small or large probabilities.

In this paper, we initially explore whether eliminating outliers during pretraining yields overall benefits to the model. Our investigation reveals that outlier-free pretraining leads to performance degradation in LLMs when employed in full precision for tasks like transfer learning or when evaluated on sequences of varying lengths compared to the pretraining sequence. Building upon this finding, we propose a novel version of clipped softmax that normalizes sequences invariantly to their length. This newly proposed method significantly boosts the LLM's performance in full precision and demonstrates reduced sensitivity to the selection of hyperparameters, a crucial aspect in pretraining as it alleviates the need for extensive grid search. Moreover, this approach also renders outlier-free pretraining effective for causal LLMs, a capability not inherent in the original clipped softmax.

%---------------------------------------------------------
\begin{figure*}[ht]
    \centering
    \includegraphics[width=0.8\linewidth]{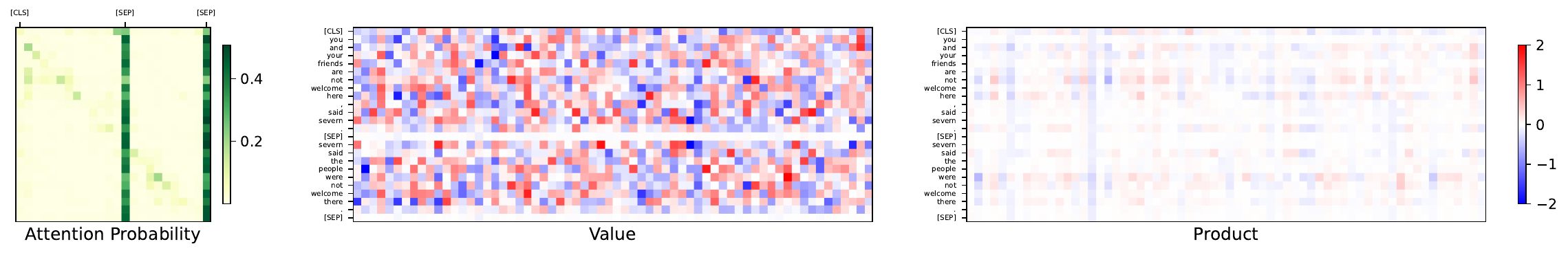}
    \caption{Common self-attention pattern of BERT-base \cite{bert}. Refer to \ref{fig: heatmap 01}, \ref{fig: heatmap 23}, \ref{fig: heatmap 45}, \ref{fig: heatmap 67}, \ref{fig: heatmap 89}, \ref{fig: heatmap 1011} for more layers.}
    \label{fig: heatmap sample}
\end{figure*}
\section{Preliminaries}
\textbf{Quantization} refers to the process of transforming high-precision values into a series of discrete levels. Our study centers on uniform affine quantization \cite{uniformQuant}, which is renowned for its optimized hardware compatibility and efficiency. The methodology employed in this research quantizes a pretrained weight or activation as follows:
\begin{align}
    \bm{X}_q = \mathrm{clip}(\nint{\frac{\bm{X}}{s}} + z, 0, 2^b-1) \label{eq: quantization}
\end{align}
where $\bm{X}$ is the weight or activation, the scale factor $s=\frac{\mathrm{max}(\bm{X})-\mathrm{min}(\bm{X})}{2^b-1}$, the zero-point $z = -\nint{\frac{\mathrm{min}(\bm{X})}{s}}$, $b$ is the bit-width, and $\nint{}$ is the round-to-nearest operation. The quantization of weight is straightforward as Equation \ref{eq: quantization}. For the quantization of activation, we calculate $s$ and $z$ with some calibration samples. During inference, one only needs to load $\bm{X}_q$ and $z$ in a reduced bit format, and $s$ in the original bit-width to GPU.

Outliers within $\bm{X}$ serve as the primary source of quantization error. They cause the scale factor $s$ to extend, resulting in fewer effective bits for most values. However, simply deleting or clipping the outliers significantly degrades the performance of quantized LLMs \cite{llmint8}. This phenomenon constitutes a central motivation behind many post-training quantization (PTQ) techniques, which aim to either maintain outliers in high precision \cite{llmint8} or empirically partition the entire tensor into smaller groups to facilitate the calculation of $s$ and $z$ \cite{awq, smoothquant, gptq}.

\textbf{Removing outliers from the source.} Instead of addressing outlier issues post-pretraining as in PTQ, recent studies \cite{qtransformer, cohere} aim to preemptively mitigate the presence of outliers during the pretraining phase, thereby rendering the pretrained model inherently more amenable to quantization.

Following conventional pretraining, a common self-attention pattern emerges, as illustrated in Figure \ref{fig: heatmap sample}, wherein certain attention heads refrain from updating the representation of certain tokens. They allocate a substantial portion of probabilities to common tokens with near-zero values in the value matrix. Consequently, the product is also near zero, contributing insignificantly to the residual connection, a phenomenon known as ``no-op'' \cite{DBLP:conf/acl/LiYYHC20, DBLP:conf/emnlp/KovalevaRRR19}.

\citet{qtransformer} posits that the ``no-op'' phenomenon primarily contributes to the presence of outliers by encouraging the concentration of attention probability on common tokens, leading to nearly zero probabilities for other tokens. According to the definition of softmax function, $\mathrm{softmax}(\bm{x}_i)=0 \Leftrightarrow \exists j \neq i, \bm{x}_j-\bm{x}_i=+\infty$. This effect is exacerbated by Layer Normalization and longer training. To address this issue, a clipped softmax (\textit{CS}) is proposed:
\begin{align}
\mathrm{clip}((\zeta - \gamma)\cdot\mathrm{softmax}(\bm{x})+\gamma, 0, 1) \label{eq: cs}
\end{align}
where $\zeta \geq 1$ and $\gamma \leq 0$. This approach aims to mitigate excessively small or large probabilities, preventing their gradient backpropagation to restrain the proliferation of outliers. CS yields improved quantizability for bidirectional encoder models, such as BERT, while offering exacerbated performance for causal LLMs like OPT \cite{opt}.

%---------------------------------------------------------

\begin{figure*}[ht]
    \centering
    \includegraphics[width=0.8\linewidth]{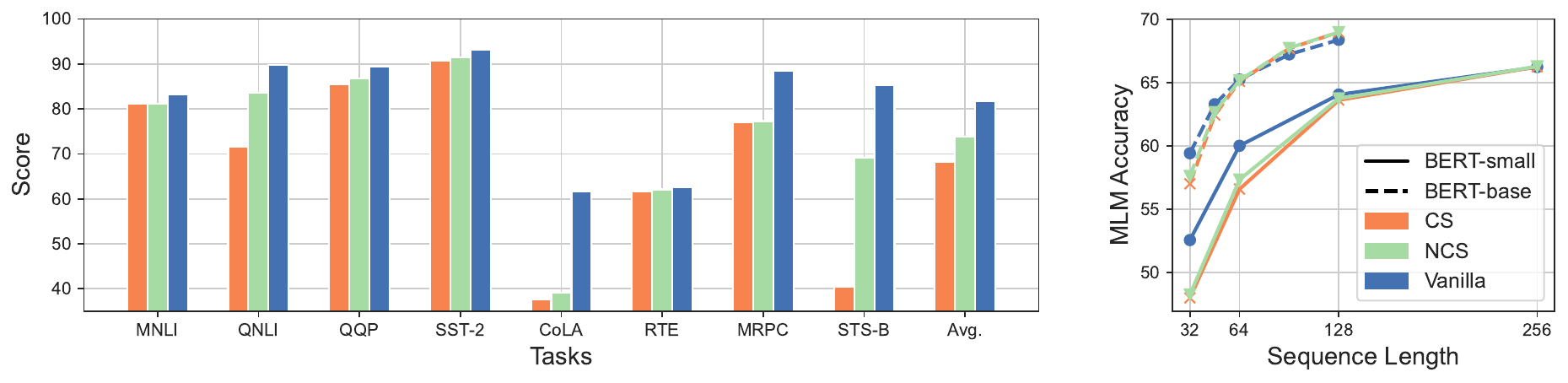}
    \caption{FP16 performance of different methods. \textbf{Left}: Finetuning results of BERT-base on GLUE. \textbf{Right}: MLM accuracy of pretrained BERT on the validation sets with different sequence lengths. BERT-small and BERT-base are pretrained with a max sequence length of 256 and 128, respectively. Refer to Table \ref{tab: glue} and \ref{tab: val length} for detailed numbers.}
    \label{fig: glue}
\end{figure*}

\begin{figure}
    \centering
    \includegraphics[width=0.7\linewidth]{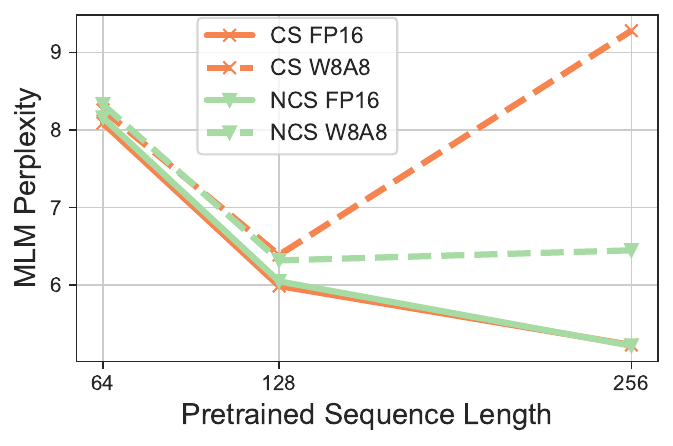}
    \caption{Pretraining BERT-small with various max sequence lengths. Refer to Table \ref{tab: pretrained length} for detailed numbers.}
    \label{fig: length}
\end{figure}

\section{Removing Outliers Is not a Free Lunch}
In this section, we delve into two key questions: (1) Is CS a panacea for BERT-like models, despite its enhancement of quantizability? (2) How can CS be tailored to effectively support causal LLMs?

\textbf{Implementation details.} We pretrain BERT \cite{bert} and OPT \cite{opt} on BookCorpus \cite{bookcorpus} and English Wikipedia, employing masked language modeling (MLM) and predicting the next token, respectively. Following pretraining, we subject the models to quantization using a stricter setting compared to most PTQ methods. This entails quantizing even the word embedding and normalization layers without partitioning tensors into smaller groups. Evaluation of BERT involves either direct measurement of its MLM accuracy/perplexity on the Wikipedia validation set or fine-tuning it on the GLUE benchmark \cite{glue}. For OPT, we evaluate its perplexity directly on the Wikipedia validation set. To gauge outliers, we additionally report the average $||\bm{X}||_\infty$ and kurtosis of $\bm{X}$ across 1024 validation samples, where $\bm{X}$ denotes activations \cite{DBLP:conf/emnlp/BondarenkoNB21}. All experiments are conducted on Transformers \cite{DBLP:conf/emnlp/WolfDSCDMCRLFDS20} with a single A6000 48GB GPU. Further details are available in Appendix $\S$\ref{sec: exp setting}.

\citet{qtransformer} conducted experiments involving an extensive search for parameters $\zeta$ and $\gamma$ in Equation \ref{eq: cs}. Optimal results were achieved with $\zeta=1$ and $\gamma=-\frac{\alpha}{T}$, where $\alpha=3.2$ for BERT, $\alpha=12$ for OPT, and $T$ represents the maximum sequence length.

\subsection{Results on BERT}
\textbf{FP16 performance of CS.} Pretraining entails significant computational demands and expenses. We contend that an ideal quantizable BERT model should also exhibit commendable performance when operating in FP16. One crucial application of pretrained BERT is its transfer learning capability to downstream tasks. Here, we finetune the pretrained BERT model on the GLUE benchmark \cite{glue} to assess its performance in FP16. As illustrated in Figure \ref{fig: glue} (left), there exists a substantial performance gap between BERT models pretrained with vanilla softmax and those with clipped softmax. Specifically, the scores of BERT models equipped with CS consistently fall below those of BERT models with vanilla softmax across all tasks, on average 68.1 versus 81.7.

However, as per the reported FP16 perplexity on the validation set, BERT equipped with CS marginally outperforms vanilla BERT, achieving scores of 4.39 versus 4.49 \cite{qtransformer}. This suggests a potential discrepancy between the pretraining and finetuning phases. We posit that differing sequence lengths encountered during pretraining and finetuning are pivotal factors contributing to this mismatch.

Let's begin by making a weak assumption, according to Equation \ref{eq: cs}, that all attention probabilities of CS fall within the range of (0, 1), implying no probabilities are clipped. In such a scenario, the normalization of Equation \ref{eq: cs} becomes:
\begin{align}
\sum_{t}^{T} \mathrm{CS}(\bm{x}_t; \zeta, \gamma) &= \zeta + (T-1)\gamma = \zeta - \alpha + \frac{\alpha}{T} \nonumber
\end{align}
Here, the normalization of CS is contingent upon the sequence length. During pretraining, all sequences maintain a consistent length for efficiency. However, in downstream tasks, the sequence lengths of samples vary, leading to distortion in the representation of the pretrained BERT.

We extend our analysis and findings, as depicted in Figure \ref{fig: glue} (right), where pretrained BERT models are directly assessed on the Wikipedia validation set. We pack the entire validation sets based on various sequence lengths and compute the MLM accuracy of pretrained BERT models. In the case of BERT-base pretrained with a maximum sequence length of 128, BERT with CS surpasses vanilla BERT when the sequence length of the validation set closely aligns with the pretraining sequence length. However, as the sequence length of the validation set decreases, BERT with CS exhibits inferior performance compared to vanilla BERT. This trend becomes more pronounced when the sequence length of the validation set significantly deviates from the pretraining sequence length, as illustrated in Figure \ref{fig: glue} (right) for BERT-small.

\textbf{Normalized Clipped Softmax (NCS).} In an effort to enhance the FP16 performance of CS, we propose a normalized variant termed NCS:
\begin{align}
\sum_{t}^{T} \mathrm{NCS}(\bm{x}_t; \zeta, \gamma) &= \zeta + (T-1)\gamma = \beta \label{eq: ncs}
\end{align}
Here, $\beta$ represents a normalization constant. The clipping mechanism remains consistent with Equation \ref{eq: cs}. We specify two hyperparameters, $\zeta$ and $\beta$. Without resorting to extensive grid search, we directly adopt optimal values from CS, setting $\zeta=1$ and $\beta=1-\alpha(T-1)/T=1-3.2(128-1)/128=-2.175$. $\gamma$ is determined by the sequence length as per Equation \ref{eq: ncs}.

One advantage of NCS lies in its robust performance across varying sequence lengths. With the normalization constant $\beta$ remaining constant, it constrains the product between attention probability and the value matrix to a range akin to pretraining. Consequently, the pretrained representation remains unaltered. This is evidenced in Figure \ref{fig: glue}, where the average score of CS improves from 68.2 to 73.8. Additionally, the MLM accuracy experiences a slight improvement. However, both the fine-tuning score and MLM accuracy still trail behind those of vanilla BERT.

\begin{table}
\begin{center}
\begin{tiny}
\begin{tabular}{c|cccc}
\toprule
\textbf{Method} & \textbf{FP16 ppl$\downarrow$} & \textbf{Max inf norm$\downarrow$} & \textbf{Avg. Kurtosis$\downarrow$} & \textbf{W8A8 ppl$\downarrow$} \\
\midrule
\multicolumn{5}{l}{\cellcolor{Gray}{\textit{BERT-small}}} \\
Vanilla & 6.17 & 801.5 & 3537.4 & 9320.60 \\
CS & 6.27 & 48.6 & 125.9 & 6.53 \\
NCS & 6.25 & 29.1 & 66.1 & 6.51 \\
\midrule
\multicolumn{5}{l}{\cellcolor{Gray}{\textit{BERT-base}}} \\
Vanilla & 4.71 & 3857.0 & 3931.9 & 4612.6 \\
CS & 4.51 & 79.0 & 160.3 & 4.93 \\
NCS & 4.51 & 79.1 & 162.8 & 4.95 \\
\midrule
\multicolumn{5}{l}{\cellcolor{Gray}{\textit{BERT-large}}} \\
Vanilla & 4.84 & 151.4 & 285.0 & 5.54 \\
CS & 4.59 & 41.2 & 25.18 & 4.78 \\
NCS & 4.61 & 43.4 & 20.4 & 4.77 \\
\midrule
\multicolumn{5}{l}{\cellcolor{Gray}{\textit{OPT-125M}}} \\
Vanilla & 15.84 & 340.0 & 1778.0 & 21.18 \\
CS & 16.29 & 63.2 & 19728.0 & 37.20 \\
NCS & 15.86 & 241.2 & 1104.5 & 18.33 \\
\bottomrule
\end{tabular}
\caption{Results on different models. NCS performs almost the same as CS on BERTs because of the same $\zeta$ and $\gamma$. \label{tab: overall}}
\end{tiny}
\end{center}
\end{table}

\textbf{Quantization performance of NCS.} A critical requirement of the pretraining method is its stability concerning the choice of hyperparameters, as grid search during pretraining is prohibitively expensive. In Figure \ref{fig: length}, we pretrain BERT-small with the default hyperparameter values of CS and NCS on training sets packed with various maximum sequence lengths. The FP16 performance of BERT-small with CS and NCS exhibits remarkable similarity. However, after quantization, NCS displays significantly lower sensitivity to the pretraining sequence length owing to normalization. This characteristic benefits pretraining and conserves computational resources by mitigating the need for hyperparameter search.

Furthermore, we explore the scalability of NCS and CS by experimenting with different scales of BERT models, an aspect missing in \citet{qtransformer}. As depicted in Table \ref{tab: overall}, both NCS and CS significantly outperform vanilla softmax on quantized BERT-small and BERT-base models. Surprisingly, quantized BERT-large with vanilla softmax performs closer to CS and NCS, contrary to the common observation that larger causal LLMs tend to exhibit more outliers \cite{llmint8}.

In conclusion, removing outliers during pretraining BERT does not offer a one-size-fits-all solution, as it may degrade its FP16 performance. NCS proves effective in enhancing the FP16 performance of CS and demonstrates reduced sensitivity to the choice of hyperparameters.

\subsection{Results on OPT}
As illustrated in Table \ref{tab: overall}, OPT-125M with CS exhibits inferior performance compared to the version with vanilla softmax for both FP16 and INT8, which contrasts with the results observed for BERT. We posit that the primary reason for this discrepancy lies in the normalization process of clipped softmax. Unlike BERT which employs bidirectional self-attention, OPT utilizes causal attention, where attention probabilities follow a lower triangular pattern, resulting in varying sequence lengths for different tokens. For instance, the sequence length of the first token is 1, while it is $T$ for the last token. CS employs a uniform sequence length $T$ for all tokens, leading to disparate normalizations of CS across different tokens, even during the pretraining stage.

The implementation of NCS for OPT slightly differs from that used for BERT whose $T$ represents the maximum sequence length. In Equation \ref{eq: ncs}, $T$ is not static for all tokens; instead, it denotes the position of the current token. Drawing inspiration from the optimal hyperparameter values of CS \cite{qtransformer}, we set $\zeta=1$ and $\beta=0.9$ for NCS. With a normalized clipped softmax wherein the sequence length varies for different tokens, the FP16 performance of NCS aligns closely with vanilla softmax, while its INT8 performance outperforms others. 

%---------------------------------------------------------
\section{Conclusion}
In this paper, we build upon the groundwork of \citet{qtransformer} by initially investigating whether the removal of outliers during the pretraining stage yields overall benefits across various aspects. Our findings reveal that this approach does not universally enhance performance, particularly concerning full precision performance. Subsequently, we introduce a normalized version of clipped softmax (NCS), aimed at enhancing the FP16 performance of BERT while also exhibiting reduced sensitivity to hyperparameter selection. Furthermore, our proposed NCS facilitates improved outlier-free pretraining of causal LLMs.

\section*{Limitations}
\textbf{Negative results.} While our main focus in this paper has been on presenting the positive outcomes of NCS, it's essential to acknowledge some negative results encountered during our study. These negative results predominantly manifest in the context of larger causal LLMs, such as OPT-350M. Despite our attempts to scale OPT-125M to OPT-350M, we observed that its performance in W8A8 quantization is significantly inferior to that of vanilla OPT-350M. This observation underscores the distinct quantization behavior of OPT compared to BERT, as evidenced in Table \ref{tab: overall}, where the $||\bm{X}||_\infty$ and kurtosis of $\bm{X}$ do not correlate well with the results of W8A8. We argue that NCS is more suited for BERT-like models and small causal LLMs.

Additionally, we note a positive outcome related to outlier-free pretraining. One noteworthy observation is that it is not necessary to train the model until convergence to evaluate its W8A8 performance. We can assess the W8A8 performance of initial checkpoints. If the gap between FP16 and W8A8 performance demonstrates an increasing trend, it indicates that the converged model may not achieve satisfactory quantization performance. This trend emerges early in the pretraining process. For instance, a significantly high perplexity of the W8A8 model may be observed after 20K iterations of BERT-base or OPT-125M, if the model is unsuitable for quantization. This trend manifests even earlier for larger models, which is advantageous for outlier-free pretraining research, as it reduces computational resource wastage.

\textbf{Further scaling experiments.} In this paper, we acknowledge that our model scaling is limited to $<350M$ parameters, which may not align with the current trends in research on large language models (LLMs). The primary constraint here is computational resources, as pretraining is a resource-intensive process. Despite this limitation, our results in Table \ref{tab: overall} demonstrate a promising trend for BERT-like models, where larger BERT models exhibit a more favorable quantization capability.

Furthermore, it's worth noting that we apply a very strict quantization setting compared to most PTQ methods, aiming to highlight the effectiveness of our proposed methods. We recognize that there is room for follow-up work to further scale the models using NCS, and we welcome future research endeavors in this direction.

\textbf{Other domains.} We acknowledge that our analysis and proposed methods have been limited to language tasks. Further evaluation of tasks in other domains, such as vision and speech, is indeed warranted. We conservatively suggest that NCS may yield better results for tasks utilizing BERT-like models, such as image recognition, semantic segmentation, and speech detection. Expanding the scope of evaluation across diverse domains will provide a more comprehensive understanding of the effectiveness and applicability of NCS beyond language tasks.

\section*{Broader Impact}
This research contributes to the field of LLM through the development of a quantization-friendly pretraining method. By facilitating more efficient deployment of LLMs, our work has the potential to democratize access to state-of-the-art AI technologies, enabling broader application across various sectors including education, healthcare, and environmental science. Such advancements can lead to significant societal benefits, such as improved educational
tools, more accurate medical diagnostics, and enhanced climate change models.

However, the ethical implications and societal consequences of the widespread implementation of quantized LLMs warrant careful consideration. The acceleration of AI capabilities could exacerbate existing disparities in technology access and literacy, potentially widening the digital divide.
Furthermore, the deployment of more efficient LLMs in sensitive areas such as surveillance, data privacy, and automated decision-making raises significant ethical concerns, including bias propagation, privacy erosion, and accountability in AI-driven outcomes.

To address these issues, it is essential for stakeholders in the AI community to engage in ongoing dialogue about the responsible development and deployment of quantized models. This includes the adoption of ethical AI frameworks, transparency in model development and deployment, and the implementation of mechanisms for accountability. Additionally, policymakers and technologists must collaborate to ensure equitable access to AI technologies, safeguard privacy, and mitigate biases in AI applications.

\bibliography{custom}

\begin{thebibliography}{20}
\expandafter\ifx\csname natexlab\endcsname\relax\def\natexlab#1{#1}\fi

\bibitem[{Ahmadian et~al.(2023)Ahmadian, Dash, Chen, Venkitesh, Gou, Blunsom, {\"{U}}st{\"{u}}n, and Hooker}]{cohere}
Arash Ahmadian, Saurabh Dash, Hongyu Chen, Bharat Venkitesh, Stephen Gou, Phil Blunsom, Ahmet {\"{U}}st{\"{u}}n, and Sara Hooker. 2023.
\newblock \href {https://doi.org/10.48550/ARXIV.2305.19268} {Intriguing properties of quantization at scale}.
\newblock \emph{CoRR}, abs/2305.19268.

\bibitem[{Bondarenko et~al.(2021)Bondarenko, Nagel, and Blankevoort}]{DBLP:conf/emnlp/BondarenkoNB21}
Yelysei Bondarenko, Markus Nagel, and Tijmen Blankevoort. 2021.
\newblock \href {https://doi.org/10.18653/V1/2021.EMNLP-MAIN.627} {Understanding and overcoming the challenges of efficient transformer quantization}.
\newblock In \emph{Proceedings of the 2021 Conference on Empirical Methods in Natural Language Processing, {EMNLP} 2021, Virtual Event / Punta Cana, Dominican Republic, 7-11 November, 2021}, pages 7947--7969. Association for Computational Linguistics.

\bibitem[{Bondarenko et~al.(2023)Bondarenko, Nagel, and Blankevoort}]{qtransformer}
Yelysei Bondarenko, Markus Nagel, and Tijmen Blankevoort. 2023.
\newblock \href {https://doi.org/10.48550/ARXIV.2306.12929} {Quantizable transformers: Removing outliers by helping attention heads do nothing}.
\newblock \emph{CoRR}, abs/2306.12929.

\bibitem[{Dettmers et~al.(2022)Dettmers, Lewis, Belkada, and Zettlemoyer}]{llmint8}
Tim Dettmers, Mike Lewis, Younes Belkada, and Luke Zettlemoyer. 2022.
\newblock \href {https://doi.org/10.48550/ARXIV.2208.07339} {Llm.int8(): 8-bit matrix multiplication for transformers at scale}.
\newblock \emph{CoRR}, abs/2208.07339.

\bibitem[{Devlin et~al.(2019)Devlin, Chang, Lee, and Toutanova}]{bert}
Jacob Devlin, Ming{-}Wei Chang, Kenton Lee, and Kristina Toutanova. 2019.
\newblock \href {https://doi.org/10.18653/V1/N19-1423} {{BERT:} pre-training of deep bidirectional transformers for language understanding}.
\newblock In \emph{Proceedings of the 2019 Conference of the North American Chapter of the Association for Computational Linguistics: Human Language Technologies, {NAACL-HLT} 2019, Minneapolis, MN, USA, June 2-7, 2019, Volume 1 (Long and Short Papers)}, pages 4171--4186. Association for Computational Linguistics.

\bibitem[{Frantar et~al.(2022)Frantar, Ashkboos, Hoefler, and Alistarh}]{gptq}
Elias Frantar, Saleh Ashkboos, Torsten Hoefler, and Dan Alistarh. 2022.
\newblock \href {https://doi.org/10.48550/ARXIV.2210.17323} {{GPTQ:} accurate post-training quantization for generative pre-trained transformers}.
\newblock \emph{CoRR}, abs/2210.17323.

\bibitem[{Jacob et~al.(2018)Jacob, Kligys, Chen, Zhu, Tang, Howard, Adam, and Kalenichenko}]{uniformQuant}
Benoit Jacob, Skirmantas Kligys, Bo~Chen, Menglong Zhu, Matthew Tang, Andrew~G. Howard, Hartwig Adam, and Dmitry Kalenichenko. 2018.
\newblock \href {https://doi.org/10.1109/CVPR.2018.00286} {Quantization and training of neural networks for efficient integer-arithmetic-only inference}.
\newblock In \emph{2018 {IEEE} Conference on Computer Vision and Pattern Recognition, {CVPR} 2018, Salt Lake City, UT, USA, June 18-22, 2018}, pages 2704--2713. Computer Vision Foundation / {IEEE} Computer Society.

\bibitem[{Jiang et~al.(2023)Jiang, Sablayrolles, Mensch, Bamford, Chaplot, de~Las~Casas, Bressand, Lengyel, Lample, Saulnier, Lavaud, Lachaux, Stock, Scao, Lavril, Wang, Lacroix, and Sayed}]{mistral}
Albert~Q. Jiang, Alexandre Sablayrolles, Arthur Mensch, Chris Bamford, Devendra~Singh Chaplot, Diego de~Las~Casas, Florian Bressand, Gianna Lengyel, Guillaume Lample, Lucile Saulnier, L{\'{e}}lio~Renard Lavaud, Marie{-}Anne Lachaux, Pierre Stock, Teven~Le Scao, Thibaut Lavril, Thomas Wang, Timoth{\'{e}}e Lacroix, and William~El Sayed. 2023.
\newblock \href {https://doi.org/10.48550/ARXIV.2310.06825} {Mistral 7b}.
\newblock \emph{CoRR}, abs/2310.06825.

\bibitem[{Kovaleva et~al.(2019)Kovaleva, Romanov, Rogers, and Rumshisky}]{DBLP:conf/emnlp/KovalevaRRR19}
Olga Kovaleva, Alexey Romanov, Anna Rogers, and Anna Rumshisky. 2019.
\newblock \href {https://doi.org/10.18653/V1/D19-1445} {Revealing the dark secrets of {BERT}}.
\newblock In \emph{Proceedings of the 2019 Conference on Empirical Methods in Natural Language Processing and the 9th International Joint Conference on Natural Language Processing, {EMNLP-IJCNLP} 2019, Hong Kong, China, November 3-7, 2019}, pages 4364--4373. Association for Computational Linguistics.

\bibitem[{Krishnamoorthi(2018)}]{DBLP:journals/corr/abs-1806-08342}
Raghuraman Krishnamoorthi. 2018.
\newblock \href {http://arxiv.org/abs/1806.08342} {Quantizing deep convolutional networks for efficient inference: {A} whitepaper}.
\newblock \emph{CoRR}, abs/1806.08342.

\bibitem[{Li et~al.(2020)Li, Yatskar, Yin, Hsieh, and Chang}]{DBLP:conf/acl/LiYYHC20}
Liunian~Harold Li, Mark Yatskar, Da~Yin, Cho{-}Jui Hsieh, and Kai{-}Wei Chang. 2020.
\newblock \href {https://doi.org/10.18653/V1/2020.ACL-MAIN.469} {What does {BERT} with vision look at?}
\newblock In \emph{Proceedings of the 58th Annual Meeting of the Association for Computational Linguistics, {ACL} 2020, Online, July 5-10, 2020}, pages 5265--5275. Association for Computational Linguistics.

\bibitem[{Lin et~al.(2023)Lin, Tang, Tang, Yang, Dang, and Han}]{awq}
Ji~Lin, Jiaming Tang, Haotian Tang, Shang Yang, Xingyu Dang, and Song Han. 2023.
\newblock \href {https://doi.org/10.48550/ARXIV.2306.00978} {{AWQ:} activation-aware weight quantization for {LLM} compression and acceleration}.
\newblock \emph{CoRR}, abs/2306.00978.

\bibitem[{Liu et~al.(2019)Liu, Ott, Goyal, Du, Joshi, Chen, Levy, Lewis, Zettlemoyer, and Stoyanov}]{roberta}
Yinhan Liu, Myle Ott, Naman Goyal, Jingfei Du, Mandar Joshi, Danqi Chen, Omer Levy, Mike Lewis, Luke Zettlemoyer, and Veselin Stoyanov. 2019.
\newblock \href {http://arxiv.org/abs/1907.11692} {Roberta: {A} robustly optimized {BERT} pretraining approach}.
\newblock \emph{CoRR}, abs/1907.11692.

\bibitem[{OpenAI(2023)}]{gpt4}
OpenAI. 2023.
\newblock \href {https://doi.org/10.48550/ARXIV.2303.08774} {{GPT-4} technical report}.
\newblock \emph{CoRR}, abs/2303.08774.

\bibitem[{Touvron et~al.(2023)Touvron, Lavril, Izacard, Martinet, Lachaux, Lacroix, Rozi{\`{e}}re, Goyal, Hambro, Azhar, Rodriguez, Joulin, Grave, and Lample}]{llama}
Hugo Touvron, Thibaut Lavril, Gautier Izacard, Xavier Martinet, Marie{-}Anne Lachaux, Timoth{\'{e}}e Lacroix, Baptiste Rozi{\`{e}}re, Naman Goyal, Eric Hambro, Faisal Azhar, Aur{\'{e}}lien Rodriguez, Armand Joulin, Edouard Grave, and Guillaume Lample. 2023.
\newblock \href {https://doi.org/10.48550/ARXIV.2302.13971} {Llama: Open and efficient foundation language models}.
\newblock \emph{CoRR}, abs/2302.13971.

\bibitem[{Wang et~al.(2019)Wang, Singh, Michael, Hill, Levy, and Bowman}]{glue}
Alex Wang, Amanpreet Singh, Julian Michael, Felix Hill, Omer Levy, and Samuel~R. Bowman. 2019.
\newblock \href {https://openreview.net/forum?id=rJ4km2R5t7} {{GLUE:} {A} multi-task benchmark and analysis platform for natural language understanding}.
\newblock In \emph{7th International Conference on Learning Representations, {ICLR} 2019, New Orleans, LA, USA, May 6-9, 2019}. OpenReview.net.

\bibitem[{Wolf et~al.(2020)Wolf, Debut, Sanh, Chaumond, Delangue, Moi, Cistac, Rault, Louf, Funtowicz, Davison, Shleifer, von Platen, Ma, Jernite, Plu, Xu, Scao, Gugger, Drame, Lhoest, and Rush}]{DBLP:conf/emnlp/WolfDSCDMCRLFDS20}
Thomas Wolf, Lysandre Debut, Victor Sanh, Julien Chaumond, Clement Delangue, Anthony Moi, Pierric Cistac, Tim Rault, R{\'{e}}mi Louf, Morgan Funtowicz, Joe Davison, Sam Shleifer, Patrick von Platen, Clara Ma, Yacine Jernite, Julien Plu, Canwen Xu, Teven~Le Scao, Sylvain Gugger, Mariama Drame, Quentin Lhoest, and Alexander~M. Rush. 2020.
\newblock \href {https://doi.org/10.18653/V1/2020.EMNLP-DEMOS.6} {Transformers: State-of-the-art natural language processing}.
\newblock In \emph{Proceedings of the 2020 Conference on Empirical Methods in Natural Language Processing: System Demonstrations, {EMNLP} 2020 - Demos, Online, November 16-20, 2020}, pages 38--45. Association for Computational Linguistics.

\bibitem[{Xiao et~al.(2023)Xiao, Lin, Seznec, Wu, Demouth, and Han}]{smoothquant}
Guangxuan Xiao, Ji~Lin, Micka{\"{e}}l Seznec, Hao Wu, Julien Demouth, and Song Han. 2023.
\newblock \href {https://proceedings.mlr.press/v202/xiao23c.html} {Smoothquant: Accurate and efficient post-training quantization for large language models}.
\newblock In \emph{International Conference on Machine Learning, {ICML} 2023, 23-29 July 2023, Honolulu, Hawaii, {USA}}, volume 202 of \emph{Proceedings of Machine Learning Research}, pages 38087--38099. {PMLR}.

\bibitem[{Zhang et~al.(2022)Zhang, Roller, Goyal, Artetxe, Chen, Chen, Dewan, Diab, Li, Lin, Mihaylov, Ott, Shleifer, Shuster, Simig, Koura, Sridhar, Wang, and Zettlemoyer}]{opt}
Susan Zhang, Stephen Roller, Naman Goyal, Mikel Artetxe, Moya Chen, Shuohui Chen, Christopher Dewan, Mona~T. Diab, Xian Li, Xi~Victoria Lin, Todor Mihaylov, Myle Ott, Sam Shleifer, Kurt Shuster, Daniel Simig, Punit~Singh Koura, Anjali Sridhar, Tianlu Wang, and Luke Zettlemoyer. 2022.
\newblock \href {https://doi.org/10.48550/ARXIV.2205.01068} {{OPT:} open pre-trained transformer language models}.
\newblock \emph{CoRR}, abs/2205.01068.

\bibitem[{Zhu et~al.(2015)Zhu, Kiros, Zemel, Salakhutdinov, Urtasun, Torralba, and Fidler}]{bookcorpus}
Yukun Zhu, Ryan Kiros, Richard~S. Zemel, Ruslan Salakhutdinov, Raquel Urtasun, Antonio Torralba, and Sanja Fidler. 2015.
\newblock \href {https://doi.org/10.1109/ICCV.2015.11} {Aligning books and movies: Towards story-like visual explanations by watching movies and reading books}.
\newblock In \emph{2015 {IEEE} International Conference on Computer Vision, {ICCV} 2015, Santiago, Chile, December 7-13, 2015}, pages 19--27. {IEEE} Computer Society.

\end{thebibliography}

\newpage
\appendix
%---------------------------------------------------------
\section{Experimental Setting}
\label{sec: exp setting}

\subsection{Pretraining}
\textbf{The pretraining data} involves BookCorpus\footnote{\url{https://huggingface.co/datasets/bookcorpus}} \cite{bookcorpus} and English Wikipedia. Following \citet{qtransformer}, we use a clean version of Wikipedia, the English subset of Wiki-40b\footnote{\url{https://huggingface.co/datasets/wiki40b}}. The training sets of BookCorpus and Wikipedia are concatenated and grouped to a specified maximum sequence length. The validation set of Wikipedia is used for direct evaluation of the pretrained model, like calculating the MLM accuracy or perplexity. 

\begin{table}
\begin{center}
\begin{small}
\begin{tabular}{lccc}
\toprule
 & \textbf{Small} & \textbf{Base} & \textbf{Large} \\
\midrule
\multicolumn{4}{l}{\cellcolor{Gray}{\textit{Architecture details}}} \\
Hidden Size & 512 & 768 & 1024 \\
Intermediate Size & 2048 & 3072 & 4096 \\
\#Attention Heads & 8 & 12 & 16 \\
\#Layers & 6 & 12 & 24 \\
\#Parameters & 35M & 109M & 335M \\
\midrule
\multicolumn{4}{l}{\cellcolor{Gray}{\textit{Pretraining details}}} \\
Max Seq. Length & 128 & 128 & 128 \\
MLM Prob. & 0.15 & 0.15 & 0.15 \\
Max Steps & 125K & 125K & 125K \\
Warmup Steps & 2K & 2K & 2K \\
Batch Size & 2048 & 2048 & 2048 \\
Peak LR & 7e-4 & 7e-4 & 1e-4 \\
LR Decay & Linear & Linear & Linear \\
AdamW ($\beta_1$, $\beta_2$) & (0.9, 0.98) & (0.9, 0.98) & (0.9, 0.98) \\
Weight Decay & 0.01 & 0.01 & 0.01 \\
Max Grad Norm & 1.0 & 1.0 & 1.0 \\
FP16 & True & True & True \\
\bottomrule
\end{tabular}
\caption{Architecture and pretraining details of BERT. A smaller learning rate for BERT-large stabilizes the pretraining.\label{tab: bert hyper}}
\end{small}
\end{center}
\end{table}

\textbf{BERT.} We experiment with three BERT models, BERT-small, BERT-base and BERT-large as Table \ref{tab: bert hyper}. Except for BERT-small which is mainly used for ablation studies, the other two variants strictly follow the original setting as \citet{bert}. Most of our pretraining settings stay the same as \citet{qtransformer} except for the batch size and maximum training steps. We modify them to follow the setting of RoBERTa \cite{roberta}, since we can faster pretrain the models. All pretraining details are shown in Table \ref{tab: bert hyper}. We pretrain all models, including vanilla, CS and NCS BERT, from scratch with the same setting for a fair comparison.

\begin{table}
\begin{center}
\begin{small}
\begin{tabular}{lcc}
\toprule
 & \textbf{125M} & \textbf{350M} \\
\midrule
\multicolumn{3}{l}{\cellcolor{Gray}{\textit{Architecture details}}} \\
Pre-norm & True & True \\
Hidden Size & 768 & 1024 \\
Intermediate Size & 3072 & 4096 \\
\#Attention Heads & 12 & 16 \\
\#Layers & 12 & 24 \\
Word Embedding Dimension & 768 & 768 \\
\midrule
\multicolumn{3}{l}{\cellcolor{Gray}{\textit{Pretraining details}}} \\
Max Seq. Length & 512 & 512 \\
Max Steps & 125K & 100K  \\
Warmup Steps & 2K & 2K \\
Batch Size & 192 & 256 \\
Learning Rate & 4e-4 & 4e-4 \\
AdamW ($\beta_1$, $\beta_2$) & (0.9, 0.95) & (0.9, 0.95) \\
Weight Decay & 0.1 & 0.1 \\
Max Grad Norm & 1.0 & 1.0 \\
FP16 & True & True \\
\bottomrule
\end{tabular}
\caption{Architecture and pretraining details of OPT. \label{tab: opt hyper}}
\end{small}
\end{center}
\end{table}
\textbf{OPT.} We experiment with two OPT variants, i.e. OPT-125M and OPT-350M. All architecture and pretraining settings strictly follow \citet{qtransformer}, as shown in Table \ref{tab: opt hyper}. Therefore, we don't pretrain vanilla and CS OPT from scratch, and directly borrow the results from \citet{qtransformer}.

\begin{table*}[ht]
\begin{center}
\begin{small}
\begin{tabular}{lcc}
\toprule
 & \textbf{MNLI, QNLI, QQP, SST-2} & \textbf{CoLA, RTE, MRPC, STS-B} \\
 \midrule
 Learning Rate & 1e-5 & \{1e-5, 2e-5, 3e-5\} \\
 Batch Size & 32 & \{16, 32\} \\
 Weight Decay & 0.1 & 0.1 \\
 Max Epochs & 10 & 10 \\
 Learning Rate Decay & Linear & Linear \\
 Warmup Ratio & 0.06 & 0.06 \\
\bottomrule
\end{tabular}
\caption{Finetuning settings on GLUE. \label{tab: glue hyper}}
\end{small}
\end{center}
\end{table*}
\subsection{Evaluation}
To evaluate pretrained BERT, we either calculate the MLM accuracy/perplexity on the Wikipedia validation set, or finetune it on the GLUE benchmark \cite{glue}. The finetuning setting inspired by \citet{roberta} is shown in Table \ref{tab: glue hyper}. To evaluate pretrained OPT, we directly calculate the perplexity of the Wikipedia validation set. 

\subsection{Quantization}
The quantization setting in this paper and \citet{qtransformer} is stricter than most PTQ methods. We quantize all weights and activations, except for the final layer, while most PTQ methods \cite{awq, smoothquant, gptq} don't quantize the word embedding and normalization layers. In addition, we don't split the whole tensor into small groups, which is also different from most PTQ methods. The motivation for such a strict quantization setting is: (1) It benefits the evaluation of the proposed methods to check whether the outliers are actually removed; (2) Due to the limitation of computation resources, we can only experiment with $<1B$ models. \cite{llmint8} shows that the outliers in larger LLM ($>6B$) are more severe. With this strict quantization setting, we attempt to treat a smaller LLM as a proxy of the larger LLM. Notably, we strictly follow the setting of \citet{qtransformer}\footnote{\url{https://github.com/Qualcomm-AI-research/outlier-free-transformers}}.

\textbf{Weights.} Following \cite{qtransformer}, we use asymmetric uniform quantization. For BERT, we use min-max weight quantization for W8A8. For OPT, we use the MSE estimator W8A8.

\textbf{Activations.} Following \cite{qtransformer}, static range estimation is applied. It calculates the scale factor $s$ and zero point $z$ with the help of some calibration samples and a running min-max estimator \cite{DBLP:journals/corr/abs-1806-08342}. For BERT, we use running min-max with 0.9 momentum over 16 batches (batch size is 8) randomly sampled from the pretraining training sets. For OPT, we use 99.999\% percentiles instead of the actual min and max over 16 batches (batch size is 8).

\begin{table*}
\begin{center}
\begin{small}
\begin{tabular}{l|cccccccc|c}
\toprule
\textbf{Method} & \textbf{MNLI} & \textbf{QNLI} & \textbf{QQP} & \textbf{SST-2} & \textbf{CoLA} & \textbf{RTE} & \textbf{MRPC} & \textbf{STS-B} & \textbf{Avg.} \\
 & m/mm & Acc & Acc/F1  & Acc & Matt & Acc & Acc/F1 & Pear/Spea & \\
\midrule
Vanilla & 83.1/83.3 & 89.8 & 90.9/87.8 & 93.1 & 61.6 & 62.5 & 86.5/90.4 & 85.3/85.0 & 81.7 \\
CS & 81.0/81.3 & 71.6 & 87.3/83.6 & 90.8 & 37.6 & 61.7 & 71.6/82.4 & 41.8/39.0 & 68.2 \\
NCS & 81.0/81.3 & 83.5 & 88.7/84.9 & 91.4 & 39.2 & 62.1 & 72.3/82.3 & 68.5/69.6 & 73.8 \\
\bottomrule
\end{tabular}
\caption{FP16 finetuning performance of BERT-base on GLUE. The median of three random runs is reported. The experimental setting is listed in Table \ref{tab: glue hyper}. \label{tab: glue}}
\end{small}
\end{center}
\end{table*}

\begin{table*}
\begin{center}
\begin{small}
\begin{tabular}{l|cccccc}
\toprule
\textbf{Length} & \textbf{32} & \textbf{48} & \textbf{64} & \textbf{96} & \textbf{128} & \textbf{256} \\
\midrule
\multicolumn{7}{l}{\cellcolor{Gray}{\textit{BERT-small, pretrained sequence length: 256}}} \\
Vanilla & 52.6	& - & 60.0	& - &  64.1 & 66.2 \\
CS & 48.0	& - & 56.6	& - & 63.6	& 66.3 \\
NCS & 48.3 & - & 57.3 & - & 63.7 & 66.3 \\
\midrule
\multicolumn{7}{l}{\cellcolor{Gray}{\textit{BERT-base, pretrained sequence length: 128}}} \\
Vanilla & 59.4	& 63.3	& 65.2	& 67.2	& 68.4 & - \\
CS & 57.0 & 62.4 & 65.1 & 67.7 & 69.0 & - \\
NCS & 57.6	& 62.7	& 65.2	& 67.7	& 69.0 & - \\
\bottomrule
\end{tabular}
\caption{MLM accuracy of pretrained BERT on the validation sets with different sequence lengths. \label{tab: val length}}
\end{small}
\end{center}
\end{table*}

\begin{table*}[ht]
\begin{center}
\begin{small}
\begin{tabular}{lc|cccc}
\toprule
\textbf{Method} & \textbf{Pretrained Seq. Len.} & \textbf{FP16 ppl $\downarrow$} & \textbf{Max inf norm $\downarrow$} & \textbf{Avg. Kurtosis $\downarrow$} & \textbf{W8A8 ppl $\downarrow$} \\
\midrule
 & 64 & 8.09 & 25.2 & 46.2 & 8.26 \\
CS ($\alpha=3.2$) & 128 & 5.99 & 48.6 & 125.9 & 6.39 \\
& 256  & 5.23 & 252.3 & 1575.5 & 9.28 \\
\midrule
& 64 & 8.16 & 30.8 & 43.5 & 8.33 \\
NCS ($\zeta=1, \beta=-2.175$) & 128 & 6.05 & 29.1 & 66.1 & 6.32 \\
& 256 & 5.22 & 102.8 & 597.2 & 6.45 \\
\bottomrule
\end{tabular}
\caption{Pretrain BERT-small with various max sequence lengths. Here we only evaluate BERT-small on 1024 validation samples. \label{tab: pretrained length}}
\end{small}
\end{center}
\end{table*}

\begin{figure*}
     \centering
     \begin{subfigure}[b]{0.48\textwidth}
         \centering
         \includegraphics[width=\textwidth]{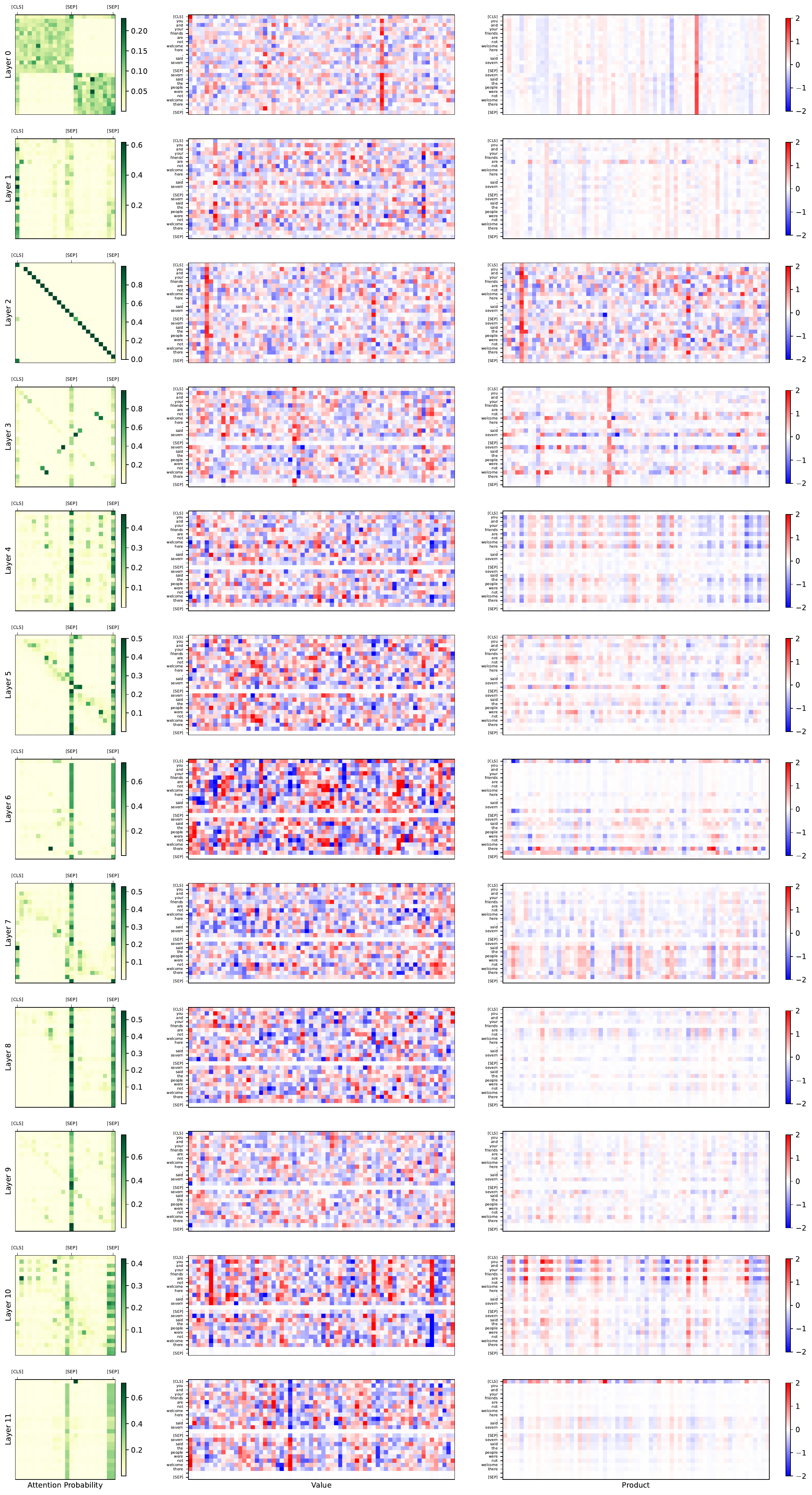}
     \end{subfigure}
     \hfill
     \begin{subfigure}[b]{0.48\textwidth}
         \centering
         \includegraphics[width=\textwidth]{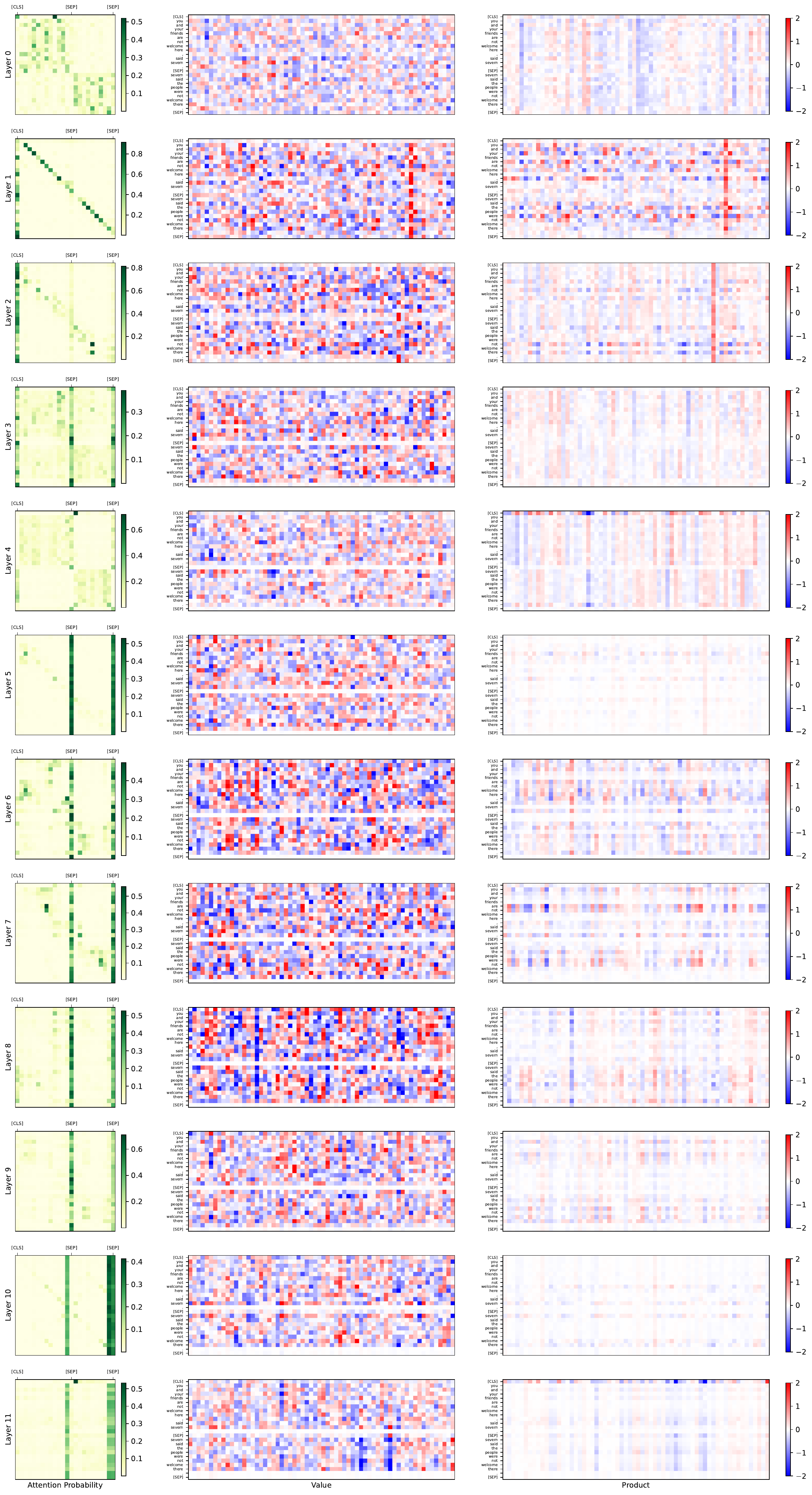}
     \end{subfigure}
     \caption{Self-attention pattern for BERT-base. \textbf{Left}: head \#0. \textbf{Right}: head \#1.}
     \label{fig: heatmap 01}
\end{figure*}

\begin{figure*}
     \centering
     \begin{subfigure}[b]{0.48\textwidth}
         \centering
         \includegraphics[width=\textwidth]{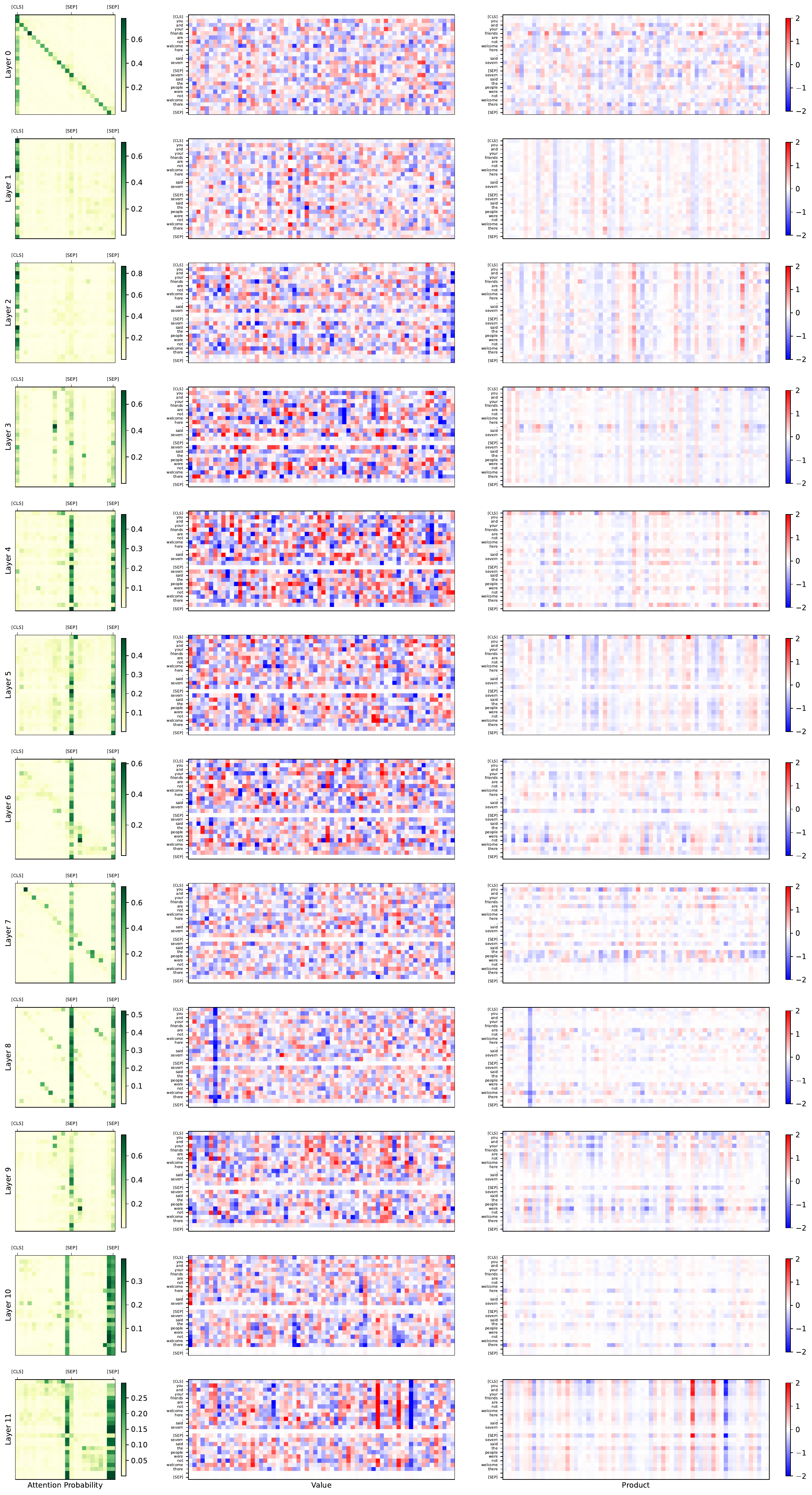}
     \end{subfigure}
     \hfill
     \begin{subfigure}[b]{0.48\textwidth}
         \centering
         \includegraphics[width=\textwidth]{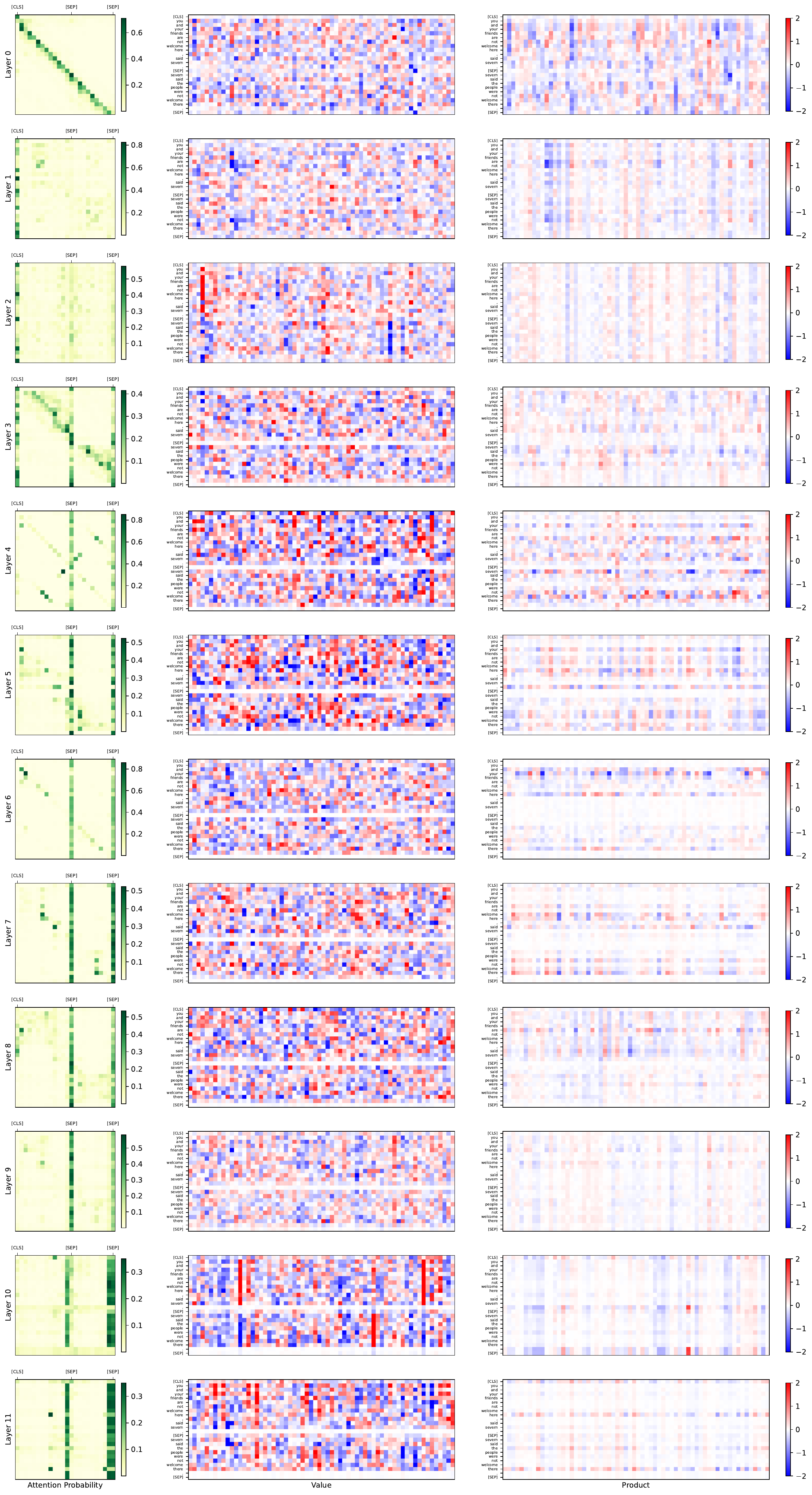}
     \end{subfigure}
     \caption{Self-attention pattern for BERT-base. \textbf{Left}: head \#2. \textbf{Right}: head \#3.}
     \label{fig: heatmap 23}
\end{figure*}

\begin{figure*}
     \centering
     \begin{subfigure}[b]{0.48\textwidth}
         \centering
         \includegraphics[width=\textwidth]{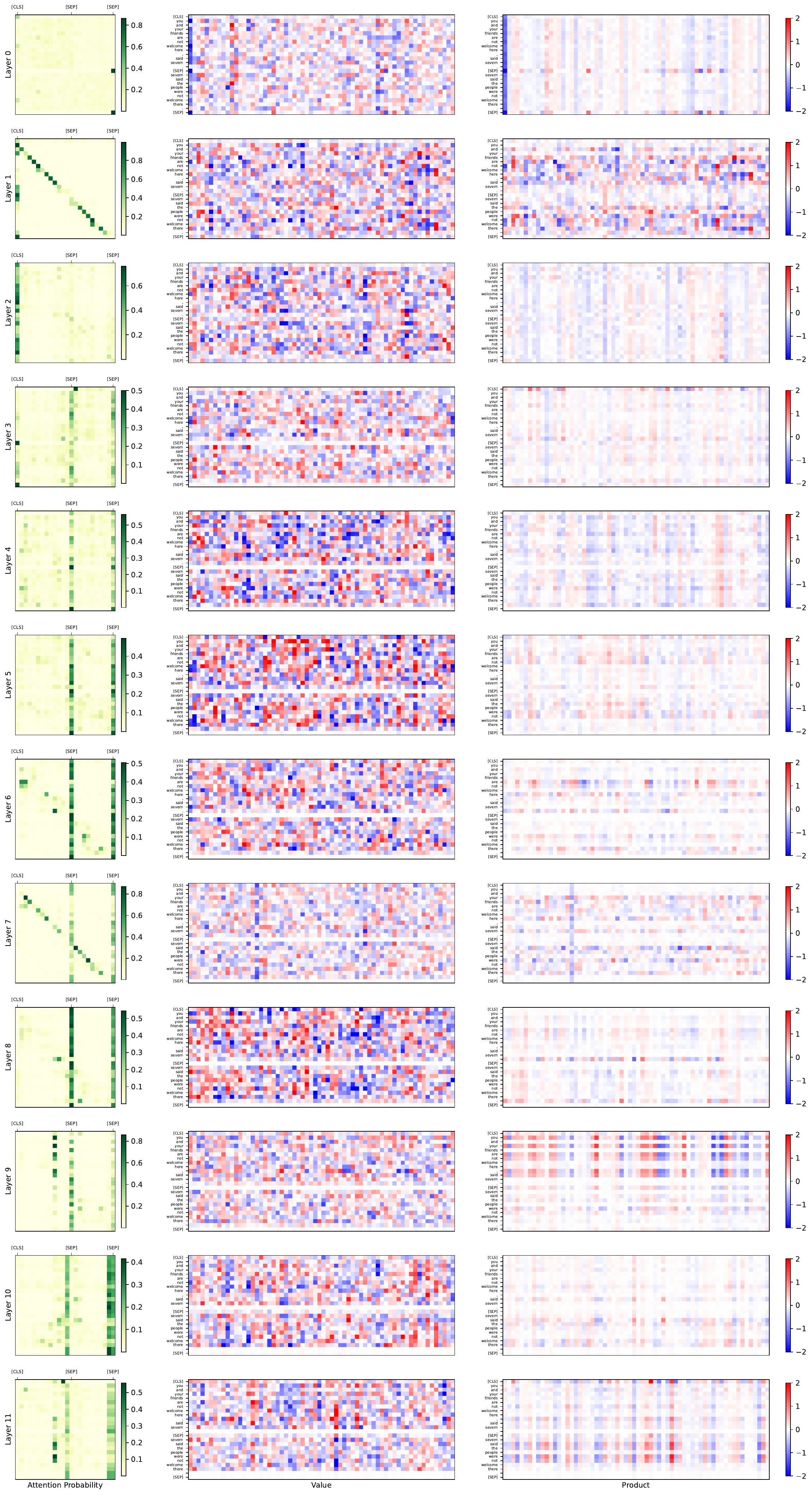}
     \end{subfigure}
     \hfill
     \begin{subfigure}[b]{0.48\textwidth}
         \centering
         \includegraphics[width=\textwidth]{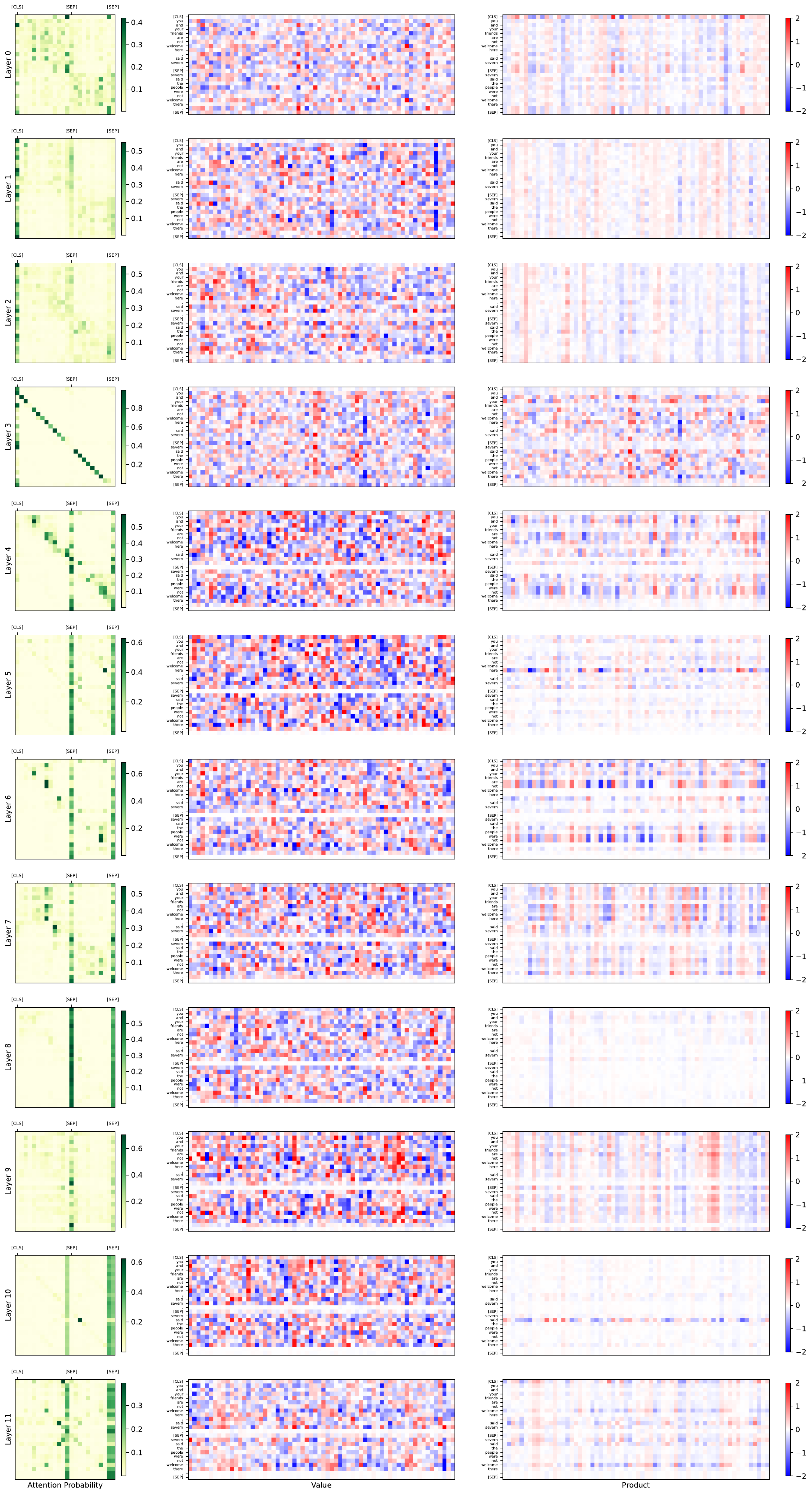}
     \end{subfigure}
     \caption{Self-attention pattern for BERT-base. \textbf{Left}: head \#4. \textbf{Right}: head \#5.}
     \label{fig: heatmap 45}
\end{figure*}

\begin{figure*}
     \centering
     \begin{subfigure}[b]{0.48\textwidth}
         \centering
         \includegraphics[width=\textwidth]{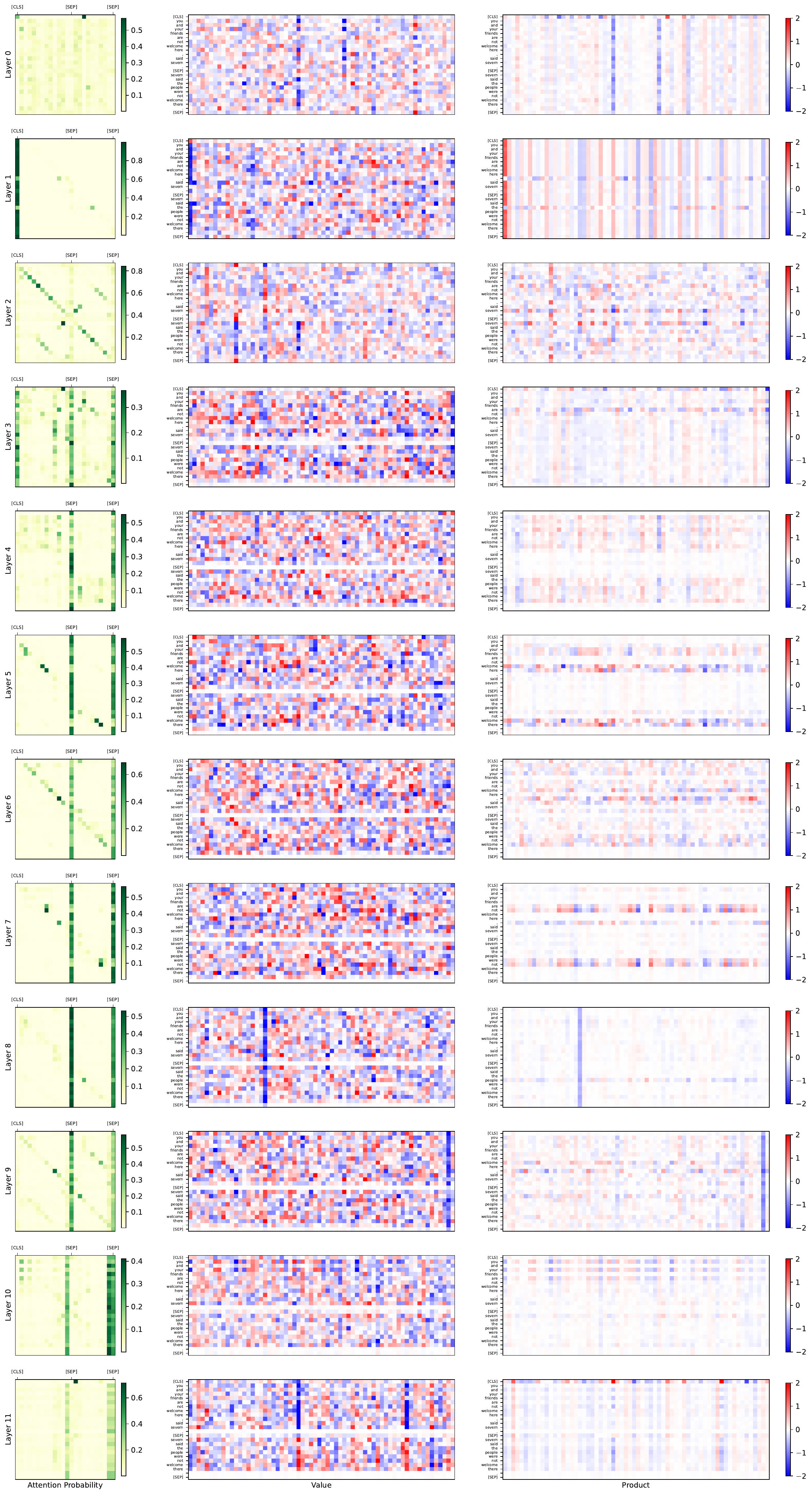}
     \end{subfigure}
     \hfill
     \begin{subfigure}[b]{0.48\textwidth}
         \centering
         \includegraphics[width=\textwidth]{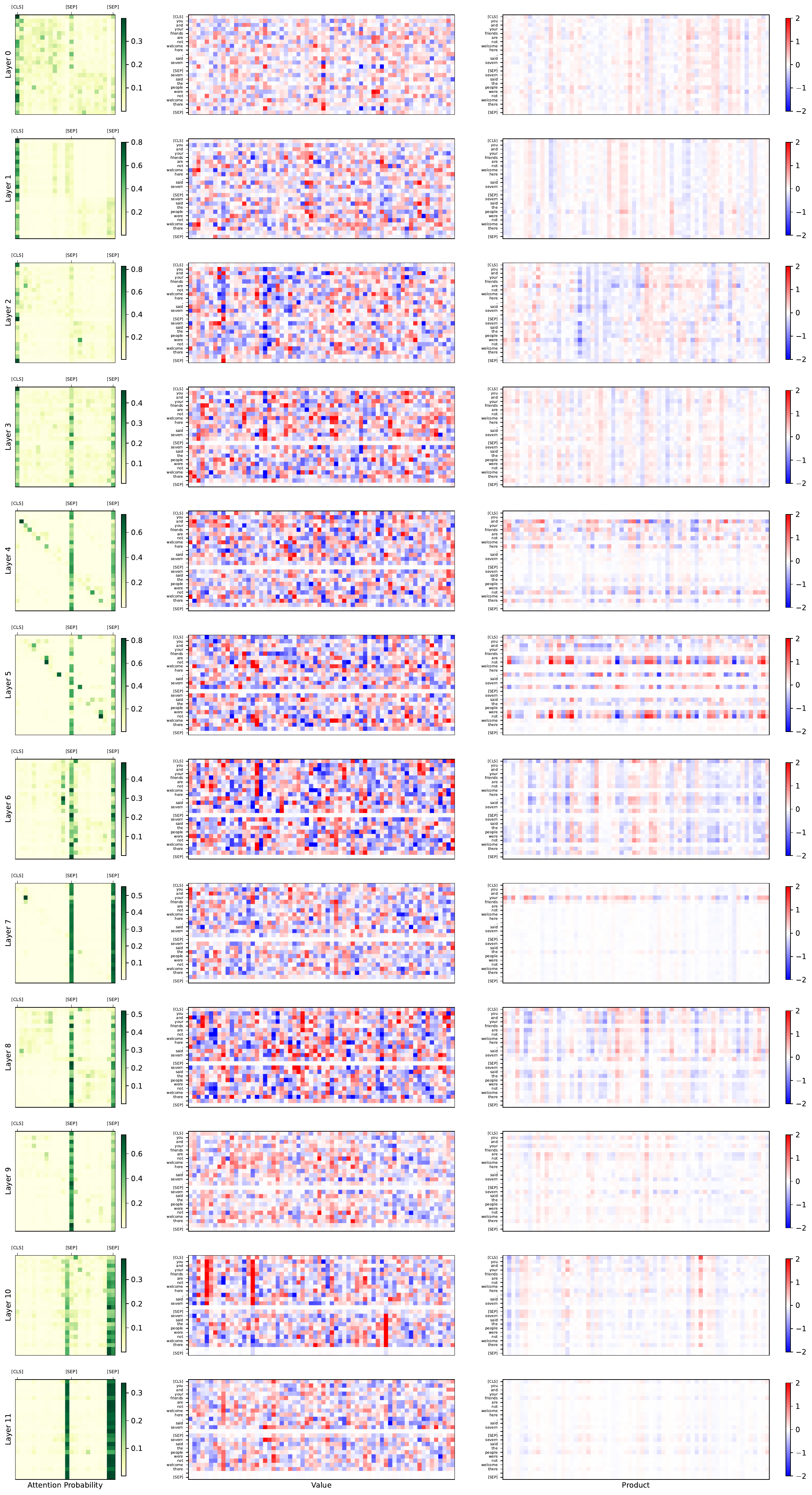}
     \end{subfigure}
     \caption{Self-attention pattern for BERT-base. \textbf{Left}: head \#6. \textbf{Right}: head \#7.}
     \label{fig: heatmap 67}
\end{figure*}

\begin{figure*}
     \centering
     \begin{subfigure}[b]{0.48\textwidth}
         \centering
         \includegraphics[width=\textwidth]{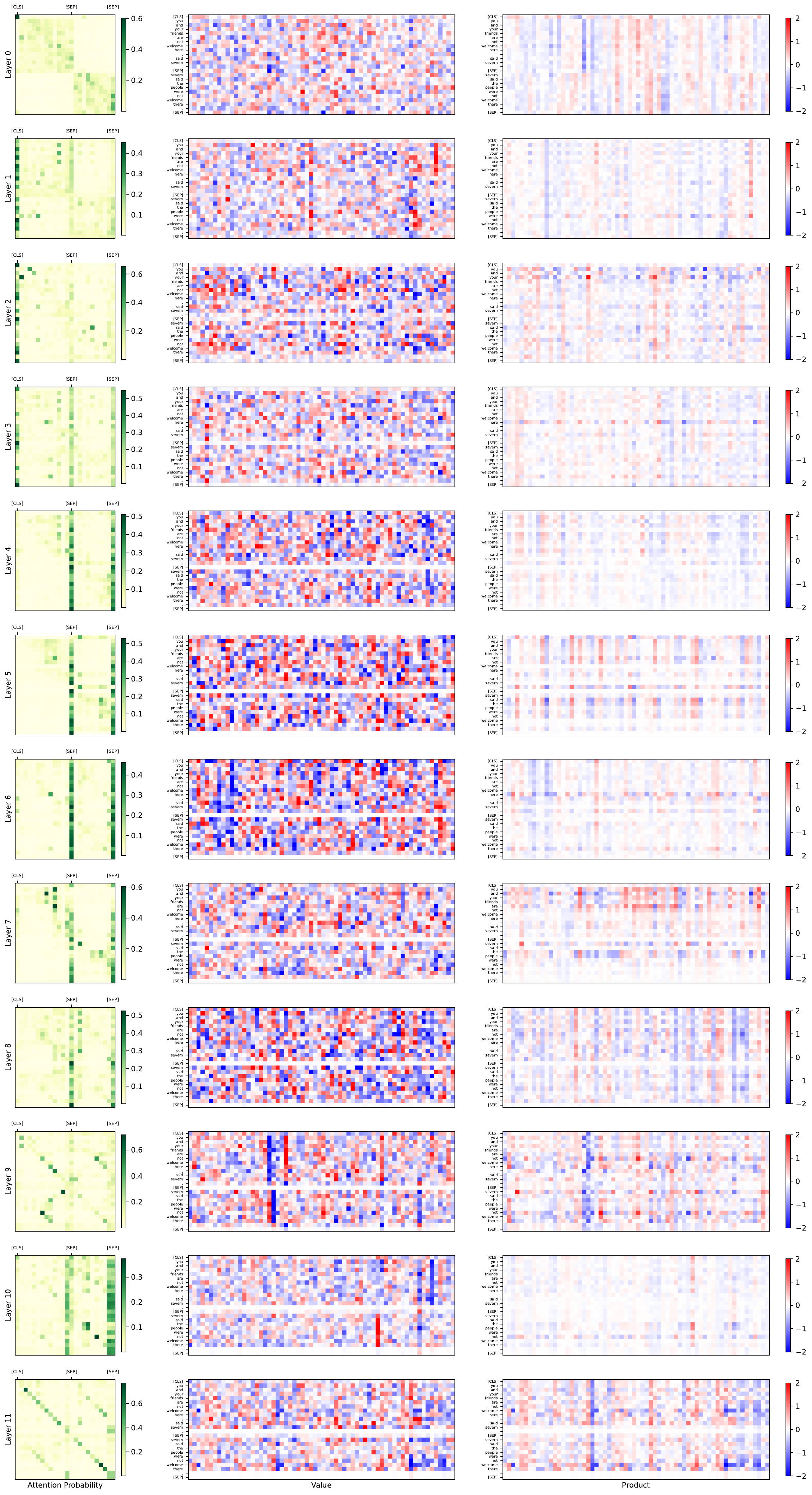}
     \end{subfigure}
     \hfill
     \begin{subfigure}[b]{0.48\textwidth}
         \centering
         \includegraphics[width=\textwidth]{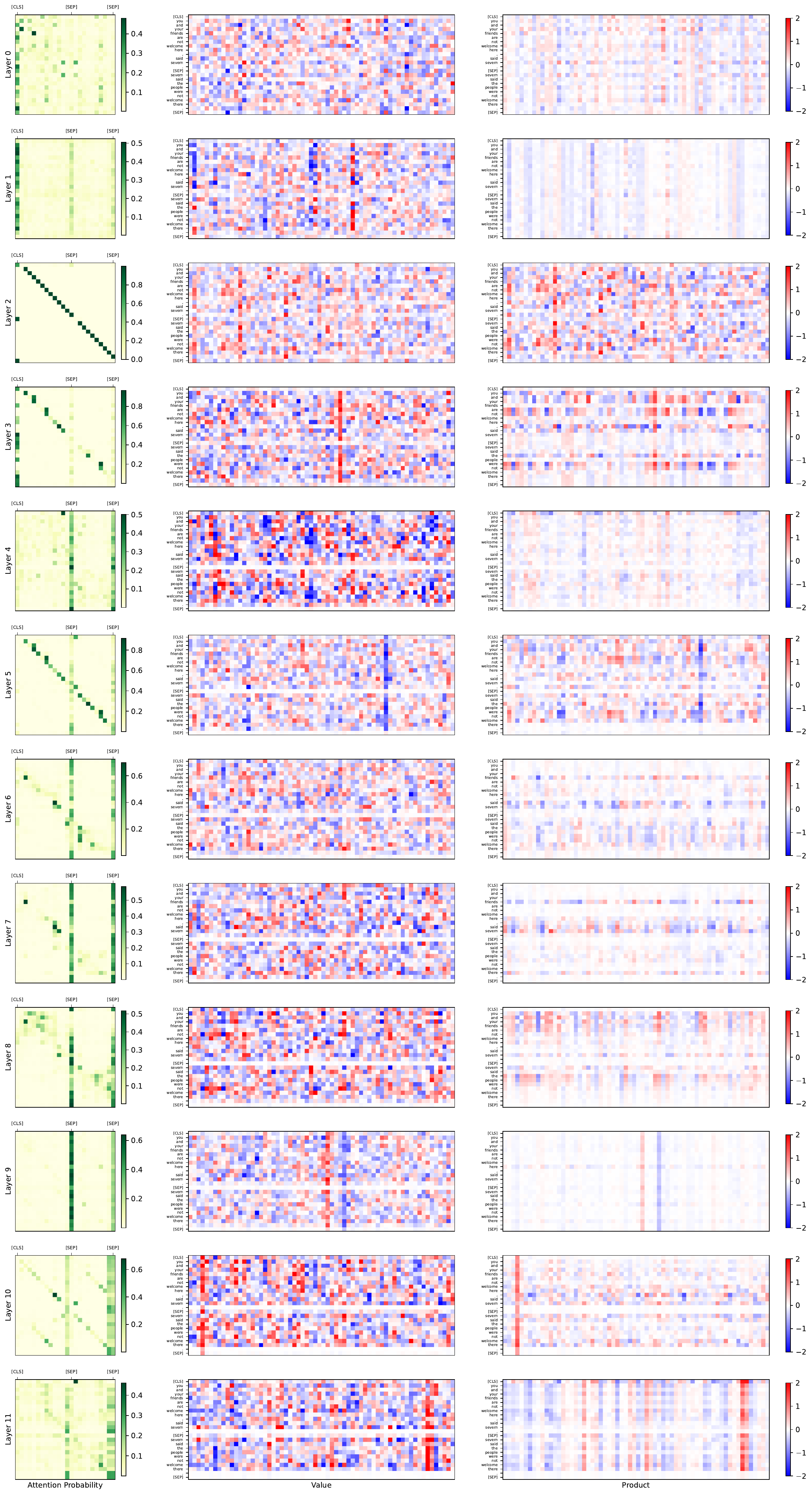}
     \end{subfigure}
     \caption{Self-attention pattern for BERT-base. \textbf{Left}: head \#8. \textbf{Right}: head \#9.}
     \label{fig: heatmap 89}
\end{figure*}

\begin{figure*}
     \centering
     \begin{subfigure}[b]{0.48\textwidth}
         \centering
         \includegraphics[width=\textwidth]{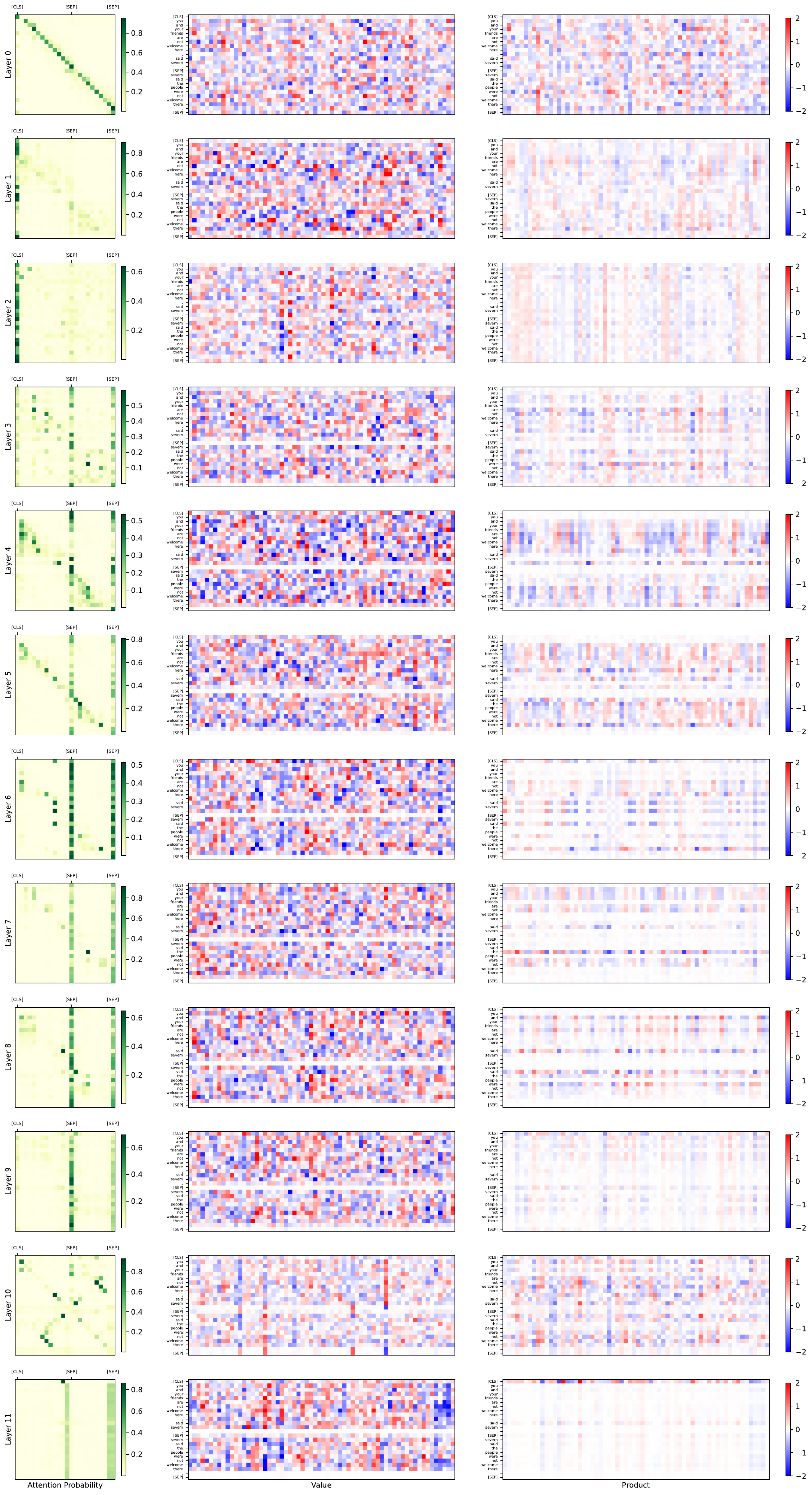}
     \end{subfigure}
     \hfill
     \begin{subfigure}[b]{0.48\textwidth}
         \centering
         \includegraphics[width=\textwidth]{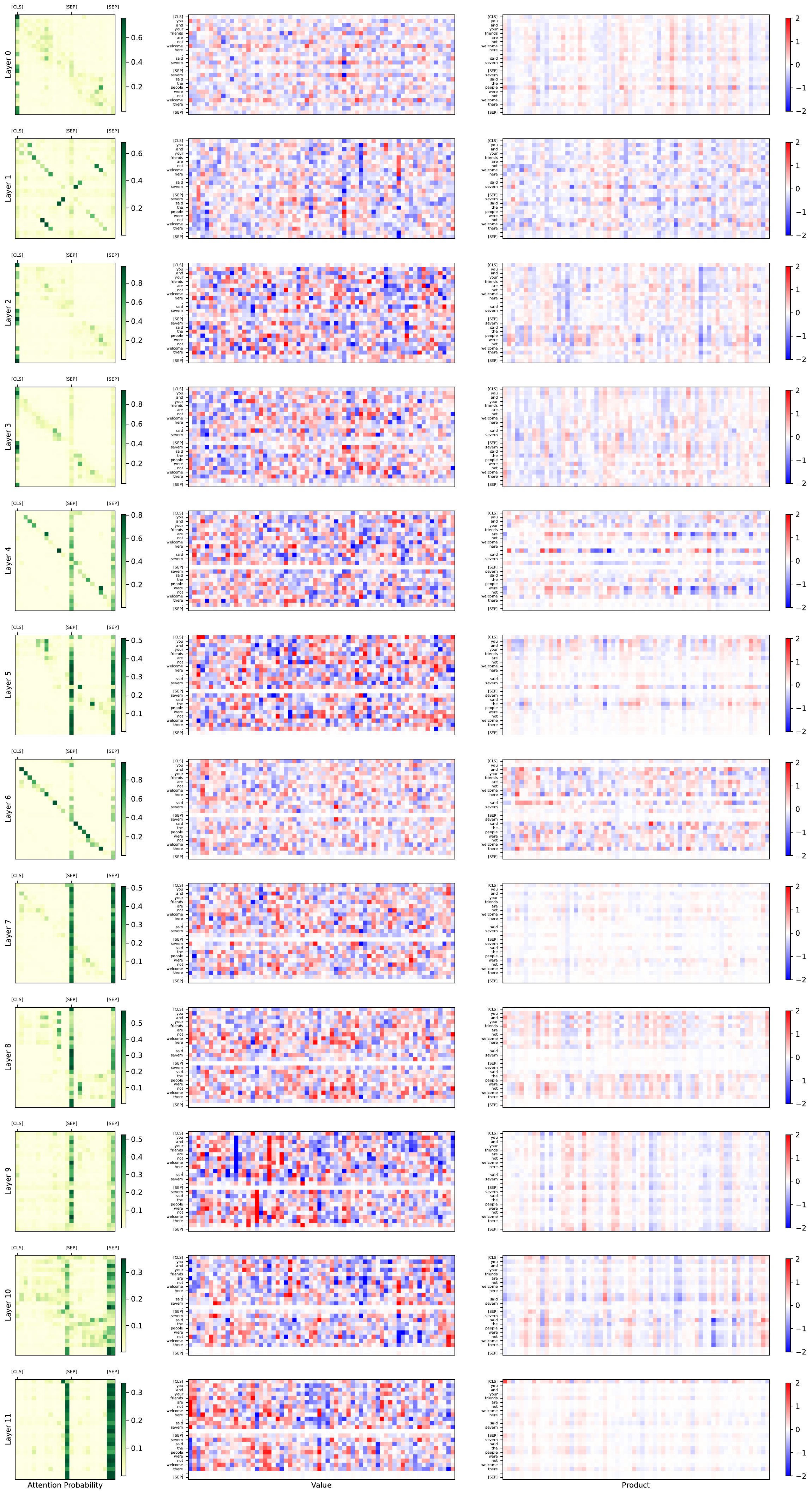}
     \end{subfigure}
    \caption{Self-attention pattern for BERT-base. \textbf{Left}: head \#10. \textbf{Right}: head \#11.}
     \label{fig: heatmap 1011}
\end{figure*}

\end{document}